\documentclass[a4paper, 10pt, conference]{ieeeconf}      

\IEEEoverridecommandlockouts                              
\usepackage{color}
\usepackage{amsfonts}
\usepackage{amsmath}
\usepackage{booktabs}
\usepackage{multirow}
\usepackage{multicol}
\usepackage{graphicx}
\usepackage{mathtools}

\newcommand{\YOON}[1]{
\textcolor{blue}{
\bfseries{YOON: {#1}}
}
}

\newcommand{\Skip}[1]{
}

\title{\LARGE \bf
Two-stream Spatiotemporal
Feature for Video QA Task}

\author{Chiwan Song, Woobin Im and Sung-eui Yoon
}

\begin{document}

\maketitle
\thispagestyle{empty}
\pagestyle{empty}

\begin{abstract}

Understanding the content of videos is one of the core techniques for developing
	various helpful applications in the real world, such as recognizing
	various human actions for surveillance systems or customer behavior
	analysis in an autonomous shop.  However, understanding the content
	or story of the video still remains a challenging problem due to its
	sheer amount of data and temporal structure.  In this paper, we
	propose a multi-channel neural network structure that adopts a
	two-stream network structure, which has been shown high performance
	in human action recognition field, and use it as a spatiotemporal
	video feature extractor for solving video question and answering
	task.  We also adopt a squeeze-and-excitation structure to
	two-stream network structure for achieving a channel-wise attended
	spatiotemporal feature.
    For jointly modeling the spatiotemporal features from video and the textual features from the question, 
    we design a context matching module with a level adjusting layer to remove the gap of information between visual and textual features by applying attention mechanism on joint modeling.
    Finally, we adopt a scoring mechanism and smoothed ranking loss objective function for selecting the correct answer from answer candidates.
    We evaluate our model with TVQA dataset, and our approach shows the
	improved result in textual only setting, but the result with visual
	feature shows the limitation and possibility of our approach.
\end{abstract}

\section{Introduction}
\Skip{In the real world, almost everything keeps changing over time
and} 
Understanding the content of videos is one of the main tasks in the computer vision
field. This is mainly because this understanding task of videos can serve as a
core technology for developing various useful applications in the real world.
For example, 
recognizing various human
actions can be useful for 
surveillance systems (e.g., detecting criminal actions) 
or customer
behavior analysis in an autonomous shop.  
Furthermore, 
understanding the story of a video also can be applied to 
assisting visually impaired people or the interactive educational system for
children.


Unfortunately, understanding the content or story of the video itself still
remains a challenging problem because of its very nature.  Since the video
consists of a sequence of images, it has a temporal structure that does not
exist in a single image. 
As a result, techniques targeting the video understanding should
not only deal with the challenges in the spatial domain such as background
clutter or object occlusion but also deal with the challenges in the temporal
domain such as moving viewpoint or reasoning the order of different events.


Thanks to the progress of deep learning technique, there have been several
deep learning based approaches for video related tasks such as video question
and answer (QA)~\cite{na2017read, Kim2017DeepStoryVS, lei2018tvqa,
gao2018motion, tapaswi2016movieqa, zhu2017uncovering, jang2017tgif,
xu2017video, mun2017marioQA}, 
or human action
recognition~\cite{simonyan2014two, carreira2017quo, wang2016temporal}.  The
approaches for solving the video QA task generally adopts
LSTM~\cite{hochreiter1997long}, GRU~\cite{chung2014empirical}
or C3D~\cite{tran2015learning},
as a spatiotemporal video feature extractor
that gets a sequence of RGB images as input.
On the other hand, approaches for solving human action recognition adopt
a two-stream ConvNet structure~\cite{simonyan2014two}, which gets RGB and
optical-flow as input to its spatiotemporal video feature extractor.
Interestingly, 
it has been known that for human action recognition
\cite{simonyan2014two, wang2016temporal, carreira2017quo},
the two-stream ConvNet spatiotemporal extractor shows
higher accuracy than those of the
single-stream ConvNet or recurrent neural network (RNN) spatiotemporal
extractor.

Inspired by the advances in this field of human action recognition, we propose
a multi-channel neural network structure with two-stream 
I3D~\cite{carreira2017quo} spatiotemporal
feature extractor for solving the video QA task.  We also adopt the
Squeeze-and-Excitation(SE) structure~\cite{hu2018squeeze} to the two-stream
I3D to apply a channel-wise attention mechanism and make the network
concentrate on important objects and actions 
in the
video.  To jointly modeling the spatiotemporal features from video and textual
features from the question, we also design a level adjusting layer to
remove the gap of information levels between two of them.
We adopt a scoring mechanism for selecting the correct
answer from answer candidates and use a smoothed ranking loss
LSEP~\cite{li2017improving} as an objective function.

We evaluate our model with the TVQA dataset~\cite{lei2018tvqa}, which provides
the sequence of video frames with subtitle as context. The dataset also
provides a question and its answer
candidates as the query.    Our approach shows the
improved result in a textual only model, but the result with the visual model shows
the limitation and possibility simultaneously.

\section{Related Work}
\begin{table*}[t]
\begin{center}
\begin{tabular}{p{2cm}|p{13cm}}
\hline
\textbf{Symbol} & \textbf{Description} \\ \hline \hline
$F_x$ & The sequence of video frames. $x$ is a type of video frames (e.g. RGB or flow) \\ \hline
$V, V^{S}, V_{x}, V_{x}^{S}$ & The spatiotemporal feature vectors from
two-stream I3D. 
$V$ without the subscript $x$ is a set of spatiotemporal feature
vectors. The superscript $S$ means the feature vectors with the time domain \\
\hline
$H_{x}$ & The textual feature vectors. $x$ is a type of textual information (e.g. subtitles or query) \\ \hline
$G_{x}$ & The context-aware query feature vector. $x$ is a type of query (e.g. $q$ is question and $a_{i}$ is $i$-th answer candidate) \\ \hline
$M_{x, i}$ & A joint embedded feature vector. $x$ is a type of context embedded with the query (e.g. video or text), and ${i}$ means $i$-th answer candidate \\ \hline
$\odot$ and $;$ & An element-wise product and a concatenation arithmetic \\ \hline
$d$ & A dimension of word embedding\\ \hline
$p_{x, i}$ & An answer probability score for a fused feature vector between
	the context $x$ and $i$-th answer
	candidate.
	\\ 
	\hline
\end{tabular}
\end{center}
\caption{The list of commonly appearing notations.}
\label{tab:notations}
\end{table*}

\subsection{Video QA task}

Video question and answering (QA) task is a challenging computer vision task
where a computer needs to answer questions given with input videos.
Though challenging, the task is worth studying since 
it is an effective way to evaluate 
how well a model understands the content of videos;
we can form any kinds of questions to test our QA models,
from naive ones (e.g., what, where, etc.) 
to more profound ones (e.g., how, why, etc.).

Most video QA methods have included subtitles or scripts
,as well as the visual cue
in that actors' lines in the text, are crucial to 
grasp essential information on videos.
Therefore, to solve the video QA task, 
the system needs to extract proper features 
from both visual inputs (i.e., RGB frames and optical flow) and
textual inputs (i.e., subtitle, query, and answer candidates), 
and adequately correlate those features to infer correct answers.
When compared to image QA task, 
video QA task has more challenges
in that it needs to additionally deal with 
the temporal domain of visual information
and connect each feature from different modalities temporally.
Even setting aside the multi-modality,
extracting good visual features rich in temporal information
itself is difficult and has been actively studied in the field of video recognition.

Researchers have been approached video QA task from various perspectives.
Na et al.~\cite{na2017read} and Kim
et al.~\cite{Kim2017DeepStoryVS} propose  
a deep model based on memory network architectures 
for embedding the story of videos and reasoning the correct answer.
Zhu et al.~\cite{zhu2017uncovering} adopt a GRU 
encoder-decoder 
for modeling the temporal structure of a video and apply
a scoring mechanism for choosing the correct answer.
Various techniques~\cite{jang2017tgif, xu2017video, mun2017marioQA, gao2018motion}
adopt a spatiotemporal attention mechanism to select important features from the appearance 
and motion information to solve the questions.
Also, 3D ConvNet~\cite{jang2017tgif,xu2017video,mun2017marioQA}  is
commonly used for extracting temporal features from RGB video frames. 

Many previous approaches use the ImageNet~\cite{imagenet_cvpr09} pre-trained
network for extracting spatial features and use
LSTM~\cite{hochreiter1997long}, GRU~\cite{chung2014empirical}, or
C3D~\cite{tran2015learning} for extracting temporal features from the sequence
of videos.
However, it has been shown for action recognition tasks that two-stream
method~\cite{simonyan2014two,wang2016temporal,fan2017identifying} that
utilizes optical flows for temporal cues has been more successful in terms of
video understanding than other methods.  Therefore, departing from previous
video QA work, we adopt a two-stream network structure for extracting useful
spatiotemporal features from the sequence of video frames.
\Skip{
The reason why we adopted a two-stream network structure instead of other previous methods
is \YOON{give a short reason here first, instead of linking}.
This will be elaborated in the subsequent section.
}

\subsection{Two-stream network structure}

Thanks to its strength in processing the spatiotemporal domain, the two-stream
network structure has been widely used in the action classification field. 
Simonyan et al.~\cite{simonyan2014two}
, Wang et al.~\cite{wang2016temporal}, and Fan et al.~\cite{fan2017identifying} 
use a two-stream ConvNet that gets two kinds of inputs: 
one is a single frame of a video for the spatial stream ConvNet, 
and the other is a multi-frame optical flow of the
video for the temporal stream ConvNet.

The two-stream network structure suggested by Carreira et al.~\cite{carreira2017quo} 
also gets two kinds of inputs
but takes both the sequence of RGB and optical flow, respectively.
Simonyan et al.~\cite{simonyan2014two}, Wang et al.~\cite{wang2016temporal}, and Carreira et al.~\cite{carreira2017quo} 
show higher accuracy on action recognition tasks 
over single-stream architectures or 
recurrent neural networks in dealing with the temporal domain of videos. 
Fan et al.~\cite{fan2017identifying} show its capability of 
processing spatiotemporal domain features with 
identifying a camera wearer from a third-person view camera scene.

In this work, we propose to use the two-stream ConvNet
for video QA task,
focusing on its ability
to process spatiotemporal domain features.

\subsection{Attention mechanism}
Attention mechanism has been widely used for various applications including
image search~\cite{Kim18, kalantidis2016cross, noh2017large}.
Since the queries of the video QA task generally ask about a specific object
or event at a specific timing in a story, solving the video QA task needs to
focus on the important information 
that is closely related to the queries from the story.

Seo et al.~\cite{seo2016bidirectional} present an attention flow layer that makes both context-aware query  and query-aware context vectors by computing a similarity matrix and using it as an attention mask.
Lei et al.~\cite{lei2018tvqa} adopt the attention flow layer 
as a context matching module and feed the context-aware vectors into bidirectional LSTM for jointly modeling the context and query.

Departing from the vector-wise attention methods, Hu et al.~\cite{hu2018squeeze} present a channel-wise attention method, 
Squeeze-and-Excitation (SE) structure, that can be applied to any given transformation 
$F_{tr} : X \mapsto U$, where $X \in \mathbb{R}^{H^{'} \times W^{'} \times C^{'}}, 
U \in \mathbb{R}^{H \times W \times C}$. SE block can be easily attached to the existing ConvNet models such as ResNet~\cite{he2016deep}
 or GoogLeNet~\cite{szegedy2015going}
, and the ConvNet with SE block shows a better result than ConvNet without SE block at the image classification task.

In this work, we apply the SE structure 
to our two-stream network 
for extracting features with
the channel-wise attention of the spatiotemporal domain.
We also utilize the context matching module
for matching the spatiotemporal features from the video frames and the queries.
\section{Our Approach}

\begin{figure*}[t]
    \begin{center}
        \includegraphics[width=\linewidth]{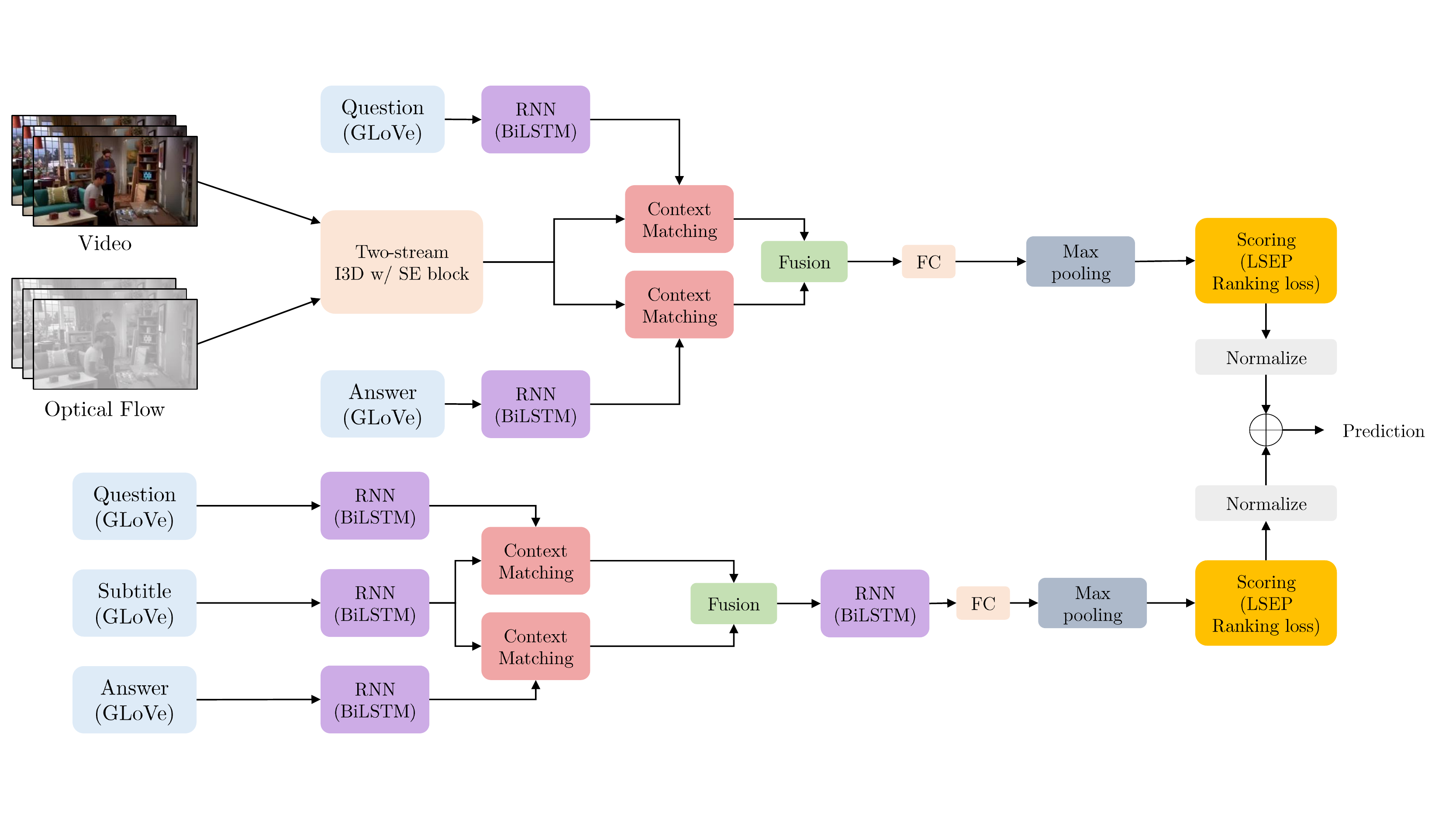}
    \end{center}
   \caption{
       The figure of our multi-channel neural network structure with two-stream spatiotemporal video feature extractor
	}
    \label{fig:overall_ex}
\end{figure*}

In this study, we propose a multi-channel neural network structure with
two-stream spatiotemporal video feature extractor for solving multimodal video
QA task. 
Our model includes a set of video and subtitle as context and
a set of question and answer candidates as the query.  Specifically, we focus on
a method for processing the video context with our novel approach, which
extracts the spatiotemporal features of videos with the channel-wise attended
two-stream network structure. 

Our approach, which is shown in Figure~\ref{fig:overall_ex},
is largely divided into two parts of the neural network stream: 
visual stream and textual stream. 
Each stream starts with corresponding feature extractors:
one for video consists of two-stream
I3D~\cite{carreira2017quo} and 
the other one for text has GLoVe~\cite{pennington2014glove} followed by bi-directional LSTM.
Both streams go through several additional layers
including context matching, fusion and scoring layer;
each stream has subtly different components and 
the details will be discussed in the next following sub-sections.
Each input (i.e. video or text) processed via
several stages produces predictions for the answer.
We aggregate the predictions at the final step
to put all information from different modalities together
and get the definitive answer.

In the next following sub-sections, we discuss the details of our approach. 
For clarity, we summarize common notations used 
throughout the paper in Table
\ref{tab:notations}. 


\subsection{Two-stream I3D with SE structure}

The visual stream of our method 
starts with the feature extraction stage.
In this work, we adopt two-stream I3D~\cite{carreira2017quo}, which shows its capacity for processing
video frames in action classification task.
However, unlike simple classification task, 
visual features for QA task are highly required to focus on
salient objects and disregard others
since it has to be correlated (i.e. context matching) to 
textual features which are 
relatively more focused on necessary context by its nature.
Therefore, we adapt Squeeze-and-Excitation (SE)~\cite{hu2018squeeze} structure for temporal inputs
and add it
to several layers of I3D extractor 
to generate more refined and attended visual features.

\begin{figure}[t]
\begin{center}
   \includegraphics[width=1.0\linewidth]{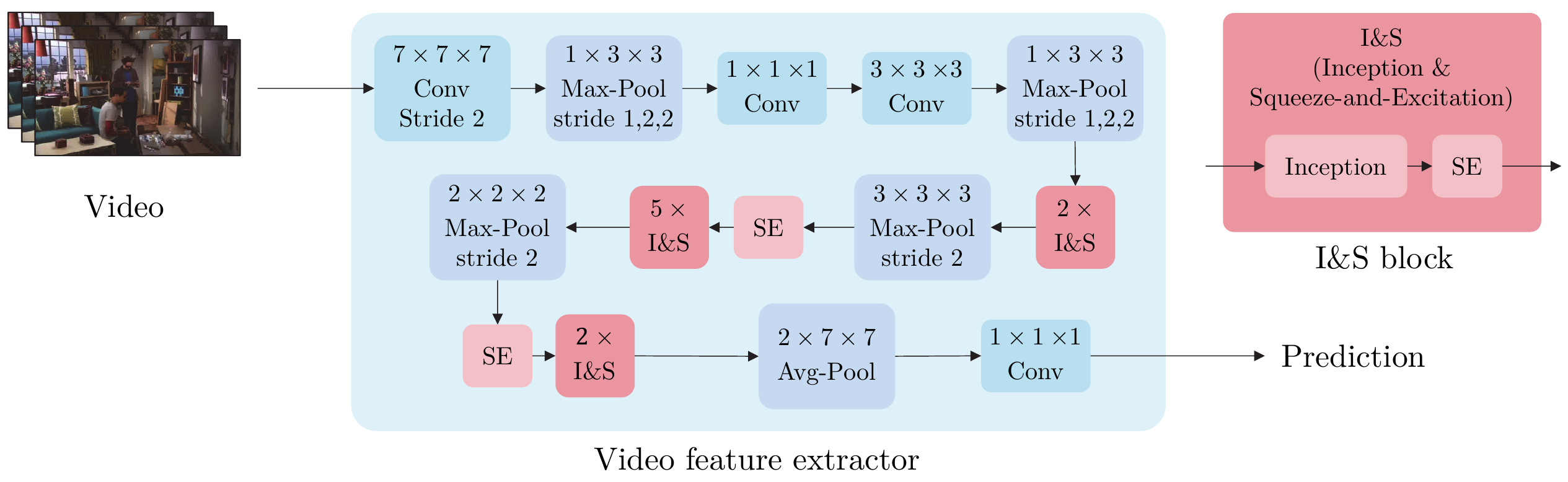}
\end{center}
   \caption{
	   Our two-stream I3D with the Squeeze-and-Excitation structure. Both
	   RGB frames and optical-flow frames are processed by video feature
	   extractor. $N$ X I\&S means $N$ different Inception \& SE
	   block
	   modules.
	}
	   \label{fig:theme_ex}
\end{figure}

Our visual feature extractor (see Fig.~\ref{fig:theme_ex}) 
is based on
I3D pretrained on ImageNet~\cite{imagenet_cvpr09} 
and Kinetics~\cite{kay2017kinetics} dataset.
It produces
a tuple of visual features 
$V^S = \{V_{spt}^{S},V_{tpr}^{S}\} $
which includes spatial feature $V_{spt}^{S}$ from 
RGB frames $F_{RGB} = \{a_0, a_1, ..., a_n\}$ and 
temporal feature $V_{tpr}^{S}$ from
flow frames $F_{flow} = \{b_0, b_1, ..., b_{n-1}\}$,
where $n$ is the number of frames in a sequence,
$a_i \in \mathbb{R}^{224\times224\times3}$
and $b_i \in \mathbb{R}^{224\times224\times2}$.
Here, we need to have temporal sequence preserved 
in extracted features
since in context matching stage, the features are
temporally matched and attended with query features.
Therefore, different from I3D~\cite{carreira2017quo}
producing two 400-dimensional vectors
from RGB and flow 
(i.e. 
$\{V_{spt}, V_{tpr} \}$, where
$V_{spt}, V_{tpr}\in\mathbb{R}^{400}$
),
we remove temporal pooling layers
to preserve the temporal sequence
so that we get
$V_{spt}^{S} \in \mathbb{R}^{n \times 400}$
and
$V_{tpr}^{S}\in\mathbb{R}^{(n-1) \times 400}$;
the preserved temporal sequence
is utilized in the context matching phase.

The video frames included in the video QA task 
commonly have unnecessary information 
(e.g. background clutters or
unrelated objects)
which degrades the performance of a video QA model.
The textual queries and subtitles, however, 
have less clutters 
because textual information is usually
well focused on related objects or actions.
This kind of gap between two modalities
raise a difficulty in context matching stage
where visual and textual information is merged.

To make our feature extractor
concentrate more on the crucial objects 
in the video frames, 
we utilize the Squeeze-and-Excitation (SE)
structure~\cite{hu2018squeeze} 
and integrate it within the two-stream I3D,
where a  
feature vector $X$ is transformed into 
another feature vector $U$ by the
inception
module~\cite{szegedy2015going, carreira2017quo}.
Fig.~\ref{fig:theme_ex} shows 
how Sqeeuze-and-Excitation structure 
is merged into the visual feature extractor.

\begin{figure}[t]
\begin{center}
   \includegraphics[width=1.0\linewidth]{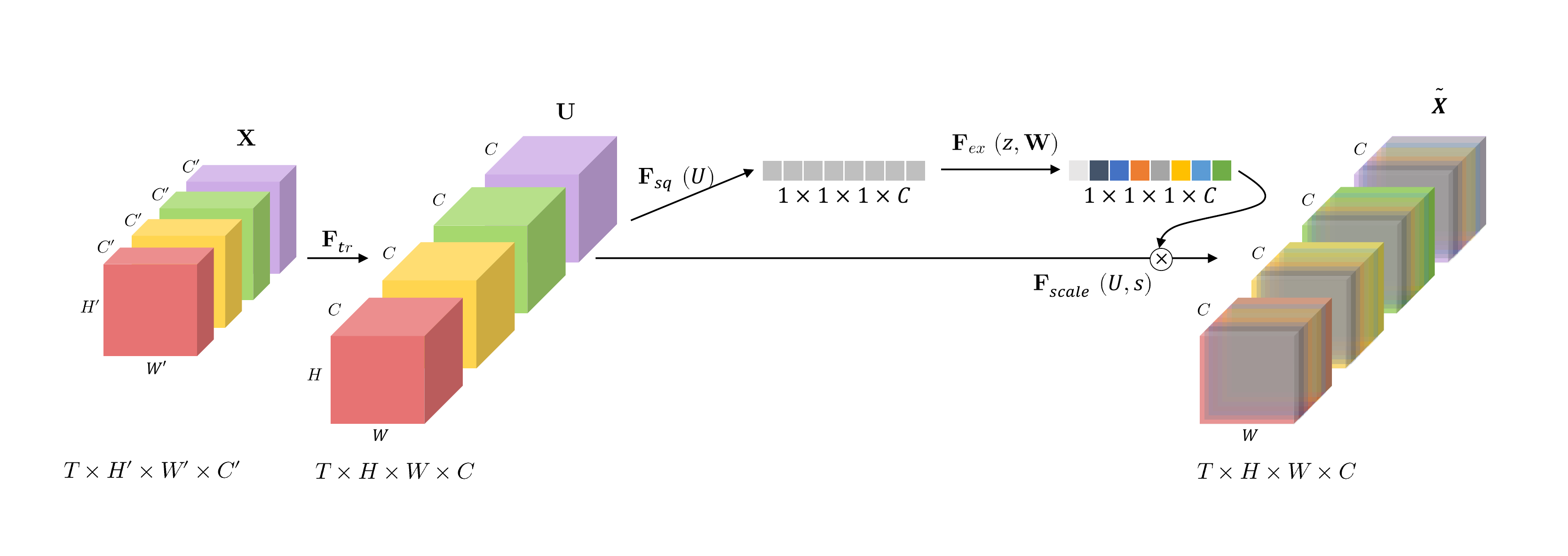}
\end{center}
   \caption{
	   Our Squeeze-and-Excitation structure 
	}
	   \label{fig:se_block}
\end{figure}

In the SE block structure of two-stream I3D, 
the \textit{Squeeze} operation
embeds global spatiotemporal information of output feature vectors of
the Inception module into a channel descriptor, 
$z \in \mathbb{R}^{1\times 1\times 1\times C}$,
with global average
pooling (Fig.~\ref{fig:se_block}).
The
\textit{Excitation} operation generates the attended channel descriptor $s$ by two
fully-connected layers with activation functions:
\begin{gather}
s=F_{ex}(z,W)=\sigma(g(z,W))=\sigma(W_2\delta(W_{1}z)),
\end{gather} 
where $z$ is the squeezed feature vectors, $\sigma$ is the sigmoid
function, $\delta$ is the ReLU~\cite{nair2010rectified} function, and
$W_1\in\mathbb{R}^{{C\over r}\times C}$ 
and $W_2\in\mathbb{R}^{C \times{C\over r}}$
are
the weight parameters of the fully-connected
dimensionality-reduction layer 
and dimensionality-increasing layer, respectively.
The FC layers include a parameter $r$ for the ratio of dimensionality reduction.

By introducing the scale function:
\begin{equation}
F_{scale}: 
\mathbb{R}^{H \times W \times L \times C}\times
\mathbb{R}^{1 \times 1 \times 1 \times C}
\rightarrow
\mathbb{R}^{H \times W \times L \times C},
\end{equation}
which operates on the channel descriptor $s$ and 
the output feature vector of the
Inception block $U$, 
we get the channel-wise attended feature vector
$\tilde{X}$:
\begin{equation}
\tilde{X} = F_{scale}(U, s) = U \odot s,
\end{equation}
where $\odot$ is the element-wise product
with shape broadcasting.
The SE block structure can be added to each inception module 
without changing the core structure of I3D
to produce better features.

\subsection{Multimodal joint embedding and context matching}

To better express the needed visual context
conditioned by queries,
extracted spatiotemporal features
should be correlated
and jointly embedded with them.
For the multimodal joint embedding,
we adopt an attention flow layer proposed by Seo et
al.~\cite{seo2016bidirectional} and Lei et al.~\cite{lei2018tvqa}.  
By the method, the spatiotemporal feature vectors and the textual feature vectors of the query are
jointly embedded to form a set of context-aware feature vectors.

We use GLoVe~\cite{pennington2014glove}
word embedding to vectorize 
each word in the query. 
Query sentences composed of multiple GLoVe vectors
are processed by bidirectional
LSTM~\cite{schuster1997bidirectional,hochreiter1997long}
which constructs our
textual features
$H_{query}\in\mathbb{R}^{n_{query} \times 2d}$, 
from the sequence of words in the query, where $n_{query}$ 
is the number of words in query consisting of the question and the answer
candidates.

Although visual features attended by SE blocks,
visual I3D feature vectors $V^{S}$ and
textual GloVe vectors $H_{query}$
have a different level of information,
and there is also innate domain gap between them. 
This gap of information level can disturb the joint embedding of
multimodal features and deliver the obscure information to the
prediction layer.

For these reasons, 
we put an information level adjusting layer to remove 
the gap of information level 
between the two types of feature vectors.
The adjusting layer consists of 
a learnable fully-connected 
layer to each feature vectors that have
$W\in\mathbb{R}^{400\times 400}$ as weight and Leaky ReLU~\cite{xu2015empirical} as an
activation function.
\Skip{For
adjusting the level of information between the two types of feature vectors, we
design a calibration layer with FC layer that have $W\in\mathbb{R}^{400}$
as a weight and Leaky ReLU~\cite{xu2015empirical} as an activation function.}
Our information-level adjusting layer produces
a set of level-adjusted
spatiotemporal feature
vectors 
$V_{spt}^{'S}\in\mathbb{R}^{n_{RGB} \times 400},
V_{tpr}^{'S}\in\mathbb{R}^{n_{flow} \times 400}$ and the calibrated textual
query feature vectors $H_{query}^{'}\in\mathbb{R}^{n_{query} \times 400}$.

The calibrated spatiotemporal feature vectors of video frames and the textual
query feature vectors are jointly modeled to produce a context-aware query
feature $G \in\mathbb{R}^{n_{video} \times 400}$ in a context matching layer
\cite{seo2016bidirectional, lei2018tvqa}:
\begin{align}
G &= S  {H'}_{query}
\\
S &= softmax(V^{'S}_{\cdots} {H'}_{query}^{T})
\end{align}
where $S \in\mathbb{R}^{n_{video} \times {n_{query}}}$ is a similarity matrix.
We calculate the similarity matrix $S$ with the matrix multiplication operation
to link each sequence of video feature vectors and query feature vectors, and
the softmax function to emphasize the important information in the similarity
matrix.

The calibrated spatiotemporal feature vectors $V^{'S}$ and the context-aware
query feature vector $G$ are fused together to form a multimodal joint embedded
feature vector $M_{video, i}=\{M_{spt}, M_{tpr}\}$:
\begin{gather}
M_{video, i} = [V^{'S};G_{q};G_{a_{i}};V^{'S} \odot G_{q};V^{'S} \odot
	G_{a_{i}}]\\M_{video, i}\in\mathbb{R}^{n_{video} \times 2000},
\end{gather}
where $n_{video}$ is the number of RGB or optical flow frames, $q$ is a
question and $a_{i}$ is the $i$-th answer candidate.
The dimension of $M_{video, i}$  is $n_{video}$ times 2,000, since we
concatenate five vectors, each of which is in 400 dimensions.

To predict the correct answer from $M_{video, i}$, we convert the fused
feature vectors $M_{video, i}$ to scalars of probability score
${p_{video,i}}=\{p_{spt}, ~p_{tpr}\}$ with an FC layer and a temporal
max-pooling layer, which can choose the most important information from the
spatiotemporal-fused feature.

\Skip{
\YOON{there have been many details here and there, but not much on the
importance or intellectual contributions. Please point them out in proper
places. Also, talk about differentiation over prior methods.}
}

\subsection{Processing of textual context and answer prediction}

We also perform the joint modeling of the query and the context encoded in another 
textual context (e.g. subtitles),
which is already included in the video QA task.
The textual context is processed by bidirectional
LSTM~\cite{schuster1997bidirectional, hochreiter1997long} and fused with the
query feature vectors to form the joint embedded feature $M_{text, i}$.
The fused feature $M_{text, i}$ is encoded with bidirectional LSTM again to
extract the temporal information and max-pooled in the temporal domain to get
the answer probability score ${p_{text,i}}$.

\Skip{Similar to the method for processing the textual query, the textual context in video QA task is embedded with GLoVe~\cite{pennington2014glove} word embedding, and encoded with the bidirectional LSTM~\cite{schuster1997bidirectional, hochreiter1997long} with $2d$ hidden states. 
The attention flow layer~\cite{seo2016bidirectional, lei2018tvqa} in Sec3.2 is also adopted for jointly embedding the features of textual context and the features of textual query, and produces the jointly embedded feature $M_{text, i}\in\mathbb{R}^{n_{text} \times 10d}$, 
where $n_{text}$ is the number of words in the corresponding textual context of QA task 
and $d$ is the number of dimension in word embedding.
The fused vector $M_{text, i}$ is converted to the} 

Finally, we normalize the answer probability score $p_{video, i}$ and
$p_{text, i}$ with the softmax function and sum up to get the final answer
probability score. 
To make the correct answer candidate's score is higher than any other wrong
answer candidates, 
we adopt the log-sum-exp pairwise (LSEP) function~\cite{li2017improving}, which
is a smooth approximation of the marginal hinge ranking loss, as an object
function:
\begin{equation}
l_{LSEP} = \log\left(1 + \sum_{v\notin Y_{i}}\sum_{u\in Y_{i}}\exp(p_{x,v}-p_{x,u}) \right),
\end{equation}
where $Y_{i}$ is the correct answer. 
By this smoothed ranking loss function, we can pose a margin between the
wrong answer candidates and the correct answer candidates in the feature space.

\section{Experiment}

\begin{table}[t]
\begin{center}
\begin{tabular}{|l|c|c|c|}
\hline
Method & S+Q & V+Q & S+V+Q \\ \hline \hline
Random & 20.00 & 20.00 & 20.00 \\ \hline 
Lei et al.~\cite{lei2018tvqa} w/ Image & \multirow{3}{*}{65.15} & 43.78 & 66.44 \\ \cline{1-1} \cline{3-4} 
Lei et al. w/ Region &  & 44.40 & 67.17 \\ \cline{1-1} \cline{3-4} 
Lei et al. w/ Concept &  & \textbf{45.03} & \textbf{67.70} \\ \hline \hline
Our Image & \multirow{4}{*}{\textbf{66.01}} & 36.00 & 64.84 \\ \cline{1-1} \cline{3-4}
Our RGB-I3D &  & 35.72 & 58.25 \\ \cline{1-1} \cline{3-4} 
Our Flow-I3D &  & 35.54 & 58.63 \\ \cline{1-1} \cline{3-4} 
Our two-stream I3D &  & 35.96 & 58.55 \\ \hline
\end{tabular}
\caption{Accuracy of different methods on TVQA validation set.
S, Q, and V indicate scripts, query, and video information, respectively.
\Skip{Lei et al. is a baseline model, which is proposed with TVQA dataset at the same time. V is the video,
S is the subtitle, and Q is the query.} 
	}
\label{tab:overall_result}
\end{center}
\end{table}

\begin{table*}[t]
\begin{center}
\begin{tabular}{|l|c|c|c|c|c|c|c|c|c|c|}
\hline
\multirow{2}{*}{} & \multicolumn{2}{c|}{S+Q} & \multicolumn{4}{c|}{V+Q} & \multicolumn{4}{c|}{S+V+Q} \\ \cline{2-11} 
 & Lei et al.~\cite{lei2018tvqa} & Ours & img & reg & cpt & Ours & img & reg & cpt & Ours \\ \hline
What(55.62\%) & 62.29 & 63.15 & 44.96 & 45.93 & 47.44 & 37.66 & 63.88 & 65.28 & 66.05 & 57.06 \\ \hline
Who(11.52\%) & 68.33 & 69.19 & 35.75 & 34.85 & 34.68 & 23.53 & 67.76 & 67.20 & 67.99 & 64.28 \\ \hline
Where(11.67\%) & 56.97 & 60.36 & 47.13 & 48.43 & 48.20 & 38.15 & 61.97 & 63.71 & 61.46 & 53.80 \\ \hline
How(8.98\%) & 71.97 & 71.10 & 41.17 & 42.41 & 40.95 & 37.69 & 71.17 & 70.80 & 71.53 & 60.96 \\ \hline
Why(10.38\%) & 78.65 & 79.14 & 45.36 & 45.36 & 45.48 & 37.73 & 78.33 & 77.13 & 78.77 & 63.23 \\ \hline
Others(1.80\%) & 74.45 & 69.92 & 36.50 & 33.58 & 33.58 & 37.15 & 73.72 & 72.63 & 74.09 & 61.72 \\ \hline
\end{tabular}
\caption{Accuracy of each question type when using baseline approaches and our
	approaches.  Ours indicate two-stream I3D feature shown in
	Table~\ref{tab:overall_result}.
	img is ImageNet features, reg is Faster-RCNN feature, cpt is the object
	detection label of Faster-RCNN, respectively.
	\Skip{V is the video, S is the subtitle, Q is the query,
	img is ImageNet features, reg is Faster-RCNN feature, cpt is the object
	detection label of Faster-RCNN, and ours is spatiotemporal features
	extracted from Two-stream I3D with SE structure}}
\label{tab:6w-result}
\end{center}
\end{table*}

\begin{table}[t]
\begin{center}
\begin{tabular}{|l|c|c|}
\hline
Method & V+Q & S+V+Q\\ \hline \hline
RGB-I3D w/o SE & 35.44 & 53.16 \\ \hline
Flow-I3D w/o SE & 35.78 & 54.16 \\ \hline
Two-stream I3D w/o SE & 35.85 & 53.91 \\ \hline \hline
RGB-I3D w/ SE & 35.72 & 58.25 \\ \hline
Flow-I3D w/ SE & 35.54 & \textbf{58.63} \\ \hline
Two-stream I3D w/ SE & \textbf{35.96} & 58.55\\ \hline
\end{tabular}
\caption{Comparison of accuracy of our methods w/ and w/o the SE block structure.
\Skip{S is the subtitle, V is the video, and Q is the
query.}}
\label{tab:se_result}
\end{center}
\end{table}

In this section, we evaluate the effectiveness of our two-stream spatiotemporal
video feature extractor in a multi-channel neural network structure.  We tested
our model on the TVQA dataset~\cite{lei2018tvqa} and compared our method
against several baseline approaches. Finally, we analyzed the prediction
accuracy of our model on different types of questions in the validation set.

\subsection{Dataset}
The TVQA dataset~\cite{lei2018tvqa} includes 152,545 QA pairs from 21,793
TV show clips.  The QA pairs are split into the ratio of 8:1:1 
for training, validation, and test sets.
The TVQA dataset provides the sequence of video frames extracted at 3 FPS, the
corresponding subtitles with the video clips, and the query consisting  of a
question and four answer candidates.  Among the four answer candidates, there
is only one correct answer.

Since the TVQA dataset provides only the sequence of RGB video frames for
the visual context, we computed the optical flow frames with TV-L1
algorithm~\cite{wedel2009improved} for our two-stream spatiotemporal feature
extractor.
The dataset also provides the timestamps for each query, so we trained and
tested our model with the timestamps to localize the video and subtitle data.

\subsection{Implementation details}


We train the textual context processing channel and the visual context
processing channel separately for reducing the variances of neural networks and
achieving the effectiveness of the model
ensemble~\cite{dietterich2000ensemble}.
\Skip{By doing this,  we can validate how
prediction result from the visual context processing channel affect the final
result,
which is a summation of the $p_{text, i}$ and $p_{video, i}$.}
When training the textual channel, we use the Adam optimizer~\cite{kingma2014adam}, where an initial learning rate is 0.0003, a momentum parameter $\beta_1$ is 0.9, a momentum parameter $\beta_1$ is 0.999, numerical stability parameter $\epsilon$ is $1e^{-8}$, and an exponential decay rate for 0.9 at every five epochs.
The model is trained for 100 epoch with the early stopping
method~\cite{caruana2001overfitting}, where the patience value is three, for
preventing the overfitting problem.  We train our textual channel on a
machine, which has Intel Xeon CPU E5-2650 v4 @ 2.20GHz, 64GB of RAM, and four
Nvidia GTX 1080Ti GPU.  The mini-batch sizes of each GPU are set to 32, and we take five days for training.

When training the visual channel, we import the pre-trained two-stream
I3D~\cite{carreira2017quo}, which was trained with the Kinetics
dataset~\cite{kay2017kinetics} that includes 25fps videos, and then fine-tune it
with the TVQA dataset. We use two kinds of optimizers, which are Adam~\cite{kingma2014adam} and SGD~\cite{bottou2010large}, 
because the two-stream I3D feature extractor
empirically requires a higher learning rate~\cite{carreira2017quo}
than other layers (e.g. context matching layer or calibration layer) in the
visual channel.  We set the maximum number of frames to 69 per question due to 
the limitation of VRAM in our GPU.  We use SGD
optimizer~\cite{bottou2010large} for
training the two-stream I3D spatiotemporal feature extractor, where an initial
learning rate is 0.02, a momentum parameter for 0.9, and an exponential decay
rate for 0.9 at every five epochs.  Each stream of two-stream I3D is trained
separately, and their predictions are  combined 
at the inference time.  

For training the calibration layer and scoring layer in the visual channel, we
use the Adam optimizer, where an initial learning rate is 0.0003, a momentum
parameter $\beta_1$ is 0.9, a momentum parameter $\beta_1$ is 0.999, numerical
stability parameter $\epsilon$ is 0.1, and an exponential decay rate for 0.9 at
every five epochs.  The model is trained for 40 epochs with the early stopping
method. A machine which we train our visual channel with has Intel Xeon CPU
E5-2630 v4 @ 2.20GHz, 1TB of RAM, and eight Nvidia RTX 2080Ti GPU.  The
mini-bach sizes of each GPU are set to 4 and it takes three weeks for training.
Our model is implemented with Tensorflow~\cite{abadi2016tensorflow}.

\subsection{Experiment results on TVQA dataset}

Table~\ref{tab:overall_result} shows the results from baseline methods to our
model on the experiment of TVQA dataset.
All experiments are tested with the timestamp option in the dataset,
which is used for localizing the video and subtitle related to the query.

We test three baseline results adopted from the work of Lei et
al.~\cite{lei2018tvqa}, which are tested with three kinds of video features:
image indicates the ImageNet~\cite{imagenet_cvpr09} pre-trained
ResNet101~\cite{he2016deep} features, which are extracted from convolutional
layer5 after pooling and has 2048 dimension, region uses the pre-trained
Faster-RCNN~\cite{anderson2018bottom, krishna2017visual} features, and concept
uses the detected object labels of the pre-trained Faster-RCNN.

Our Image, RGB-I3D, Flow-I3D, and Two-stream I3D shown in Table~\ref{tab:overall_result} are variations of our model:
Image uses the pre-trained ResNet101 features, whose dimension is reduced to
400 for fitting to our model,
RGB-I3D uses the spatial video features from the sequence of video RGB frames,
Flow-I3D uses the temporal video features from the sequence of optical-flow
frames, and Two-stream I3D uses both spatiotemporal video feature.  The random
selection model gets 20\% accuracy because the TVQA dataset has one correct
answer among five answer candidates.

\Skip{V+Q in the second column of table means that the model predict the answer with the video and query as the source of information,
and S+V+Q in the third column of table means that the model predict the answer with the video feature, subtitle, and query as the source of information.}

When we evaluate the models with subtitle and query information, our model gets
better results than the tested baseline methods. We assume that the LSEP loss
function, which we adopt instead of cross-entropy loss, helps our model to
learn better features and solve more difficult questions than the baseline
model. 

We expected that the features from the two-stream network show higher accuracy
than using the ImageNet feature, but the ImageNet feature from ResNet101, Our Image,
gets the highest accuracy when considering both
video and query among the tested four different types of video features
evaluated with our model. 
Furthermore, the ImageNet feature in our model shows a lower result than the
baseline methods with the ImageNet feature. This is caused by the reduction of
dimension on the ImageNet feature to fit our model to the context matching,
degrading the amount of information in the feature.

Under the test setting of S+V+Q, we find that all kinds of video features show
lower accuracy than the baseline, and they even degrades the performance of S+Q
setting.  Especially, the video features from I3D much degrades the performance
of S+Q setting than the ImageNet feature (Our Image).  To find out why our
video features get an inferior result over the baseline and degrade the
accuracy of the text-only setting, we analyze the
baseline and our approaches' accuracy of each question type.

\Skip{Among the three of our video features, the spatiotemporal feature from
two-stream I3D gets the highest accuracy when considering both video and query
information.  
This is mainly because the spatial features from the RGB-I3D stream and the
temporal features from the Flow-I3D stream complement each other and
provide relatively
rich information than each single stream feature.  
However, our video features do not work well compared to the baseline and get a lower accuracy under the
test setting of S+Q.  To find out why our video features get an inferior result
over the baseline and degrade the accuracy of the text-only setting, we analyze the
baseline and our approaches' accuracy of each question type.} 


Table~\ref{tab:6w-result} shows the accuracy of different methods under each
question type. We split the questions into six types of what, who, where, how,
why, and others.
As shown in S+Q column, our text only model improves accuracy in every question
types except for 'why' and 'others' which account for 10\% of the total
questions.  Specifically, our model improves the most in 'where' questions,
where the original model shows the lowest accuracy.

On the contrary to the result of S+Q, our model shows lower accuracy than
baseline methods in every question types except for 'others' in V+Q.  Our
hypothesis of this result is that our spatiotemporal feature extractor with the
two-stream I3D has difficulty in extracting the feature from the provided video
data, because 3fps, which is a frame rate of provided video data, is too low
for extracting sufficiently dense optical-flow for our two-stream I3D, which is
originally trained on 25fps videos. Nevertheless, our V+Q model shows the
highest accuracy on 'where' question type, which our S+Q model shows the lowest
accuracy.  This result
shows the possibility that our visual model can complement the textual model.

\subsection{The effectiveness of the SE structure}

To see the effectiveness of our SE structure, we perform ablation study w/ and
w/o the SE block structure in Table~\ref{tab:se_result}.
We find that the models with the SE structure show higher accuracy than the
models without the SE structure in S+V+Q. Based on this result, we can see that
the SE structure in our method helps to
extract the complementing spatiotemporal features to the textual feature.


\section{Conclusion}
In this paper, we have proposed a multi-channel neural network structure for
solving the video QA task.  We use two-stream I3D for extracting temporal
features from the sequence of video frames, and apply the SE structure between
the Inception blocks in the two-stream I3D. 
The SE structure gives the effect of channel-wise attention to the network, 
so it makes the network to focus on the important objects in the video.
For multimodal joint embedding of the video and text features, we modify
two-stream I3D to produce the features with the sequence in it, and we also
design the information-level adjusting layer to reduce the gap of information
levels between the two types of features.

To evaluate our approaches, we conducted the experiment with TVQA dataset and
ablation study with the SE structure.  Our approach showed the improved result
in the textual model, but the result with the visual model showed its limitation
with
possible future research directions.
The ablation study with SE structure showed the effectiveness of the SE structure
that makes the visual model to complement the textual model.

\subsection{Limitations and future work}
Since the visual channel in our model spent too much time for training, we
needed more than a week to check the experiment result.  This heavy
requirement of computational power led to the insufficient amount of attempts
for finding a proper architecture design and hyper-parameters.

In future work, we will train and evaluate our model in a more efficient way
and search for the cause of the malfunction in the visual
channel, such as modifying the structure of I3D to work with low frame videos or
checking the context module whether it works properly to reduce the gap of
information level.


\bibliographystyle{ieeetr}
\bibliography{egbib}

\end{document}